\documentclass[twocolumn, switch]{article} 
\usepackage{indentfirst}
\usepackage{preprint}
\PassOptionsToPackage{hyphens}{url}\usepackage{hyperref}
\usepackage{graphicx}
\usepackage{adjustbox}
\usepackage[bottom]{footmisc}
\usepackage{amsmath, amsthm, amssymb, amsfonts}

\usepackage[numbers,square]{natbib}
\usepackage[utf8]{inputenc}	
\usepackage[T1]{fontenc}	
\usepackage{xcolor}		
\usepackage{booktabs} 		
\usepackage{nicefrac}		
\usepackage{microtype}		
\usepackage{lineno}		
\usepackage{float}			
\usepackage{newfloat}
\DeclareFloatingEnvironment[name={Supplementary Figure}]{suppfigure}
\usepackage{sidecap}
\sidecaptionvpos{figure}{c}

\usepackage{titlesec}
\titlespacing\section{0pt}{8pt plus 3pt minus 3pt}{1pt plus 1pt minus 1pt}
\titlespacing\subsection{0pt}{6pt plus 3pt minus 3pt}{1pt plus 1pt minus 1pt}
\titlespacing\subsubsection{0pt}{4pt plus 3pt minus 3pt}{1pt plus 1pt minus 1pt}

\title{Improving the Deployment of Recycling Classification through Efficient Hyper-Parameter Analysis}

\usepackage{xwatermark}

\usepackage{authblk}

\begin{document}
\author{
\begin{tabular}[t]{c@{\extracolsep{15em}}c} 
Mazin Abdulmahmood \& Ryan Grammenos \\
mazin.abdulmahmood.18@ucl.ac.uk     r.grammenos@ucl.ac.uk \\
Electrical and Electronic Engineering   \\ 
University College London  \\
\end{tabular}
}

\twocolumn[ 
  \begin{@twocolumnfalse} 
  
\maketitle
\begin{abstract}
\textbf{The paradigm of automated waste classification has recently seen a shift in the domain of interest from conventional image processing techniques to powerful computer vision algorithms known as convolutional neural networks (CNN). Historically, CNNs have demonstrated a strong dependency on powerful hardware for real-time classification, yet the need for deployment on weaker embedded devices is greater than ever. The work in this paper proposes a methodology for reconstructing and tuning conventional image classification models, using EfficientNets, to decrease their parameterisation with no trade-off in model accuracy and develops a pipeline through TensorRT for accelerating such models to run at real-time on an NVIDIA Jetson Nano embedded device. The train-deployment discrepancy, relating how poor data augmentation leads to a discrepancy in model accuracy between training and deployment, is often neglected in many papers and thus the work is extended by analysing and evaluating the impact real world perturbations had on model accuracy once deployed. The scope of the work concerns developing a more efficient variant of WasteNet, a collaborative recycling classification model. The newly developed model scores a test-set accuracy of 95.8\% with a real world accuracy of 95\%, a 14\% increase over the original. Our acceleration pipeline boosted model throughput by 750\% to 24 inferences per second on the Jetson Nano and real-time latency of the system was verified through servomotor latency analysis.}
\end{abstract}
\vspace{0.35cm}

  \end{@twocolumnfalse} 
] 



\section{Introduction}
Recent years have seen tremendous leaps in performance for machine learning algorithms in a wide range of typically human-based activities \cite{one}. Progression in the field has led to ever increasing complexity and depth of developed models. Embedded machine learning seeks to bring machine learning (ML) to the edge for real world deployment and plays a vital role in shifting deep learning (DL) to more realistic and practical scenarios. Typically, ML algorithms are computer-intensive tasks that require powerful processing or when the target device is weaker, external processing executed through the cloud. When the target environment is isolated, such conditions cannot be supported and there rises the need to use embedded systems to achieve deployment. The current internet backbone is almost at capacity \cite{two} leading to increased latency when transmitting data, crippling time-sensitive applications. A device at the ‘edge’ runs all processing or tasks locally on its own hardware and limits the need for net transmissions. The key benefit is that less unnecessary data is required to be transmitted and so latency issues are alleviated. Furthermore, the demand for societal integration of artificially intelligent products grows, but the problem lies in shifting the deployment paradigm onto embedded devices as it is not commercially viable to couple all products with powerful hardware. Such shifts follow platform-independent \cite{three} or hardware-specific approaches. \par
This paper takes a hardware-specific approach which chooses to solely optimise the model’s performance on its own architecture. The board of choice is an NVIDIA Jetson Nano. The end goal is for efficient use of board resources and to meet performance goals. Performance in a hardware-specific approach can be measured by the following:
\begin{itemize}
\item Top 1 percent accuracy (before and after optimisation).
\item Throughput of the network - the volume of output within a given period for example the number inferences per second (IPS)
\item Percentage CPU/ GPU load from overhead.
\item Memory footprint of the model (MB).
\item Efficiency of the network – amount of throughput delivered per unit-power (performance/Watt).
\end{itemize}
\par
The NVIDIA Jetson Nano is a solid-state embedded device rivalling the Raspberry Pi. The Jetson Nano offers a small, portable form factor with low power consumption (a few Watts sitting idle), GPIO capabilities, a high performance-to-price ratio and is less than a tenth of the price of a desktop PC \cite{four}. The Nano is seen as a middle ground between the cheap \$35 Pi and the next more powerful Jetson family device, the TX2, at \$90 \cite{eight}, and hence is the chosen edge device for deployment.\par
The work performed continues on the work of \cite{nine} in building a waste classification model (WasteNet) that classifies recycling into five categories: Cardboard, Glass, Metal, Paper and Plastic based on the TrashNet dataset from \cite{ten}. The model achieved an accuracy of 95.40\% on the test-set. The results from \cite{nine} were applied to develop the benchmark model for comparison within this paper. The work uses real world performance analysis to improve upon WasteNet’s generalisation capabilities at deployment, as well as accelerate the model to run at real-time on the Jetson Nano. A study on audio-visual response time in \cite{twsix} identified that the visual stimuli response time in humans is around 190ms, and so to match that at a system level we classify a real-time system to be better than that of humans. 10 IPS is therefore the real-time performance criteria which requires the model to have a response time of 100ms. This paper also explores the use of EfficientNets as a base model and acts to verify the accuracy-parameterisation trade-off claims of EfficientNets presented in \cite{eleven}. The code developed for the work is published on the GitHub paper repository \cite{git} and a demonstration of real-time classification on the Jetson Nano is available at \cite{demo}.\par
This paper is organised as follows: Section 2 explores related work in the field; Section 3 introduces the system design and architecture of both the benchmark and the newly developed efficient  variants of WasteNet, as well it provides a background on TensorRT, the chosen acceleration library, and also covers latency analysis details. The subsequent sections present the methodology behind model reconstruction and acceleration and evaluates the real world performance of the model when deployed onto the Jetson Nano. The work is then summarised in Section 8 and future work is covered in Section 9.

\section{Related Work}
Performing image classification on embedded devices is no new venture, many everyday devices, such as mobile phones, advertise an image classification based facial recognition unlock feature as a selling point in hopes of drawing in consumers. In fact, as of 2018 over 1 billion smartphones are predicted to have implemented some sort of facial recognition system \cite{twelve} and this number is expected to grow to 1.3 billion by 2024 \cite{thirteen}. \par 
Often the main problem with deploying large models at the edge is that they are over-parameterised, that being during inferences only a few weights within the network layers tend to contribute to the final classification, leading to redundant memory accesses that waste energy and memory scratch space within devices \cite{fourteen}. In 2020 the authors of \cite{fifteen} looked to break down these large deep neural network (DNN) classifiers into multiple smaller DNNs called modules \cite{fifteen}. This aims to reduce computational processes by using an initial root model to select the general group of categories the image belonged to and following it by several modules that distinguish among these similar grouped categories through an averaged Softmax likelihood recursively until classification. This approach reduced memory requirements by 50\% to 99\%, inference times by 55\% to 94\% and floating-point operations (FLOPS) by 15\% to 99\%. Ensemble methods, methods that use multiple networks to classify, typically lead to training of weak classifiers \cite{sixteen} and therefore accuracy is the trade-off for throughput which is a noted result from the paper as the modular neural network scores the highest test error of 0.313 on the ImageNet dataset (for a comparison ResNet-34 has a test-error of 0.276).\par
Concerning deployment on the Nano board, many papers deploy the model as vanilla and few delve into giving acceleration configurations leaving novel developers to discover them on their own. The author of \cite{seventeen} implemented a generic object tracker on the Nano using two parallel CNNs which inferenced at 60 frames per second (FPS) on a computer but dropped drastically to 10FPS on the Jetson Nano even after pruning the model, which negatively impacted the model’s accuracy. Another author, \cite{nineteen}, compared the use of TensorRT, NVIDIA’s acceleration library, to another acceleration library known as TensorFlow Lite (TFLite) - a library that converts a TensorFlow model into a compressed flat buffer which can then be pruned and quantised \cite{eighteen}. TFLite is mainly aimed at edge devices such as mobile phones that do not have dedicated GPUs. It was found that the TFLite model inferenced faster at 0.0227s contrasted against TensorRT’s 0.0545s, however, it is important to point out this is an audio-based CNN that produced minimal load on the GPU. TFLite is designed to accelerate CPU based models whereas TesnorRT specialises in accelerating GPU based models. We can see this as the max CPU utilisation of the TFLite model sat at 33.8\%, 9.8\% lower than the TensorRT model yet it inferenced much quicker, showing that TensorRT did not handle optimisations on the CPU as well.
The NVIDIA GPU delegates found in the Jetson Nano are not supported in TFLite, further to this, TFLite optimisation of image based CNNs was evaluated in \cite{twenty} by manually binding the Nano's Tegra GPU to TFLite and the results showed that there was no beneficiary speed increase on the Tegra GPU that would make it a viable Jetson optimisation route.\par
Around the subject of waste classification, there are a few papers that incorporate real world deployment. One paper, \cite{wastenetwhite}, used the technique of transfer learning, a widely used training technique relating to training pre-trained models on a new domain, and applied this to their model through the TrashNet dataset. They also used the idea of fine-tuning the model through freezing and retraining of the model. The authors claim a test-set accuracy of 97\% but there is ambiguity surrounding specific implementation of the model architecture and training details. In this paper the same technique of transfer learning is used, however the model architecture is not based on classic CNN bases such as VGG and ResNets. Furthermore we introduce a more suitable technique for fine-tuning whereby the network layers never require freezing. Our dataset also allows us to introduce mixed-source data training for generalisation evaluation. \par
Of the papers that did cover real world deployment, the authors of \cite{twone} built a hybrid DNN that classified recycling objects. The model scored an accuracy of 98.2\% on their own test-set when the recycling object was held in a standard, fixed position but scored only 91.6\% when the objects where rotated or oriented, for example holding a bottle upside down. Perception issues seem to be a common problem when deploying models within the real world. Most papers use data augmentation techniques such as rotating training samples to mimic these orientations which aid in increasing real world accuracies \cite{twtwo}. It is apparent the skew of objects varies as they are rotated in space and tend to be of a different depth which a simple 2D image rotation in post-augmentation cannot fully replicate, which could explain the discrepancy in real world performance to training performance \cite{twthree}.\par 
The author of \cite{twfour} developed a waste classification model, using the TrashNet dataset and transfer learning on an Xception based model architecture. The paper used TFLite to compress the model and deploy it on a Raspberry Pi 4.  The model failed to run at real-time and could only singly inference through the camera module lending evidence to TFLite’s inability to accelerate GPU models. The model accuracy achieved is stated to be 92\%, however it is not apparent whether this is their test-set accuracy or peak validation accuracy achieved. This paper is one of a couple found to implement waste classification on an edge device, however it gives no detail to inferencing speeds on the Pi, nor does it give any specific acceleration recommendations.\par 
The gap in research surrounding edge waste classification is apparent. This paper funds further research into this domain and proposes a tuned approach to training and deploying highly efficient models that run at real-time with no trade-offs between speed and accuracy when accelerating and deploying models on edge devices. This paper also analyses and evaluates techniques for improving real world performance of the model and provides insight into what to expect when deploying image-based networks at the edge.

\section{WasteNet System Design}
\subsection{Overall System Design}
\begin{figure}[H]
        \includegraphics[width=0.5\textwidth]{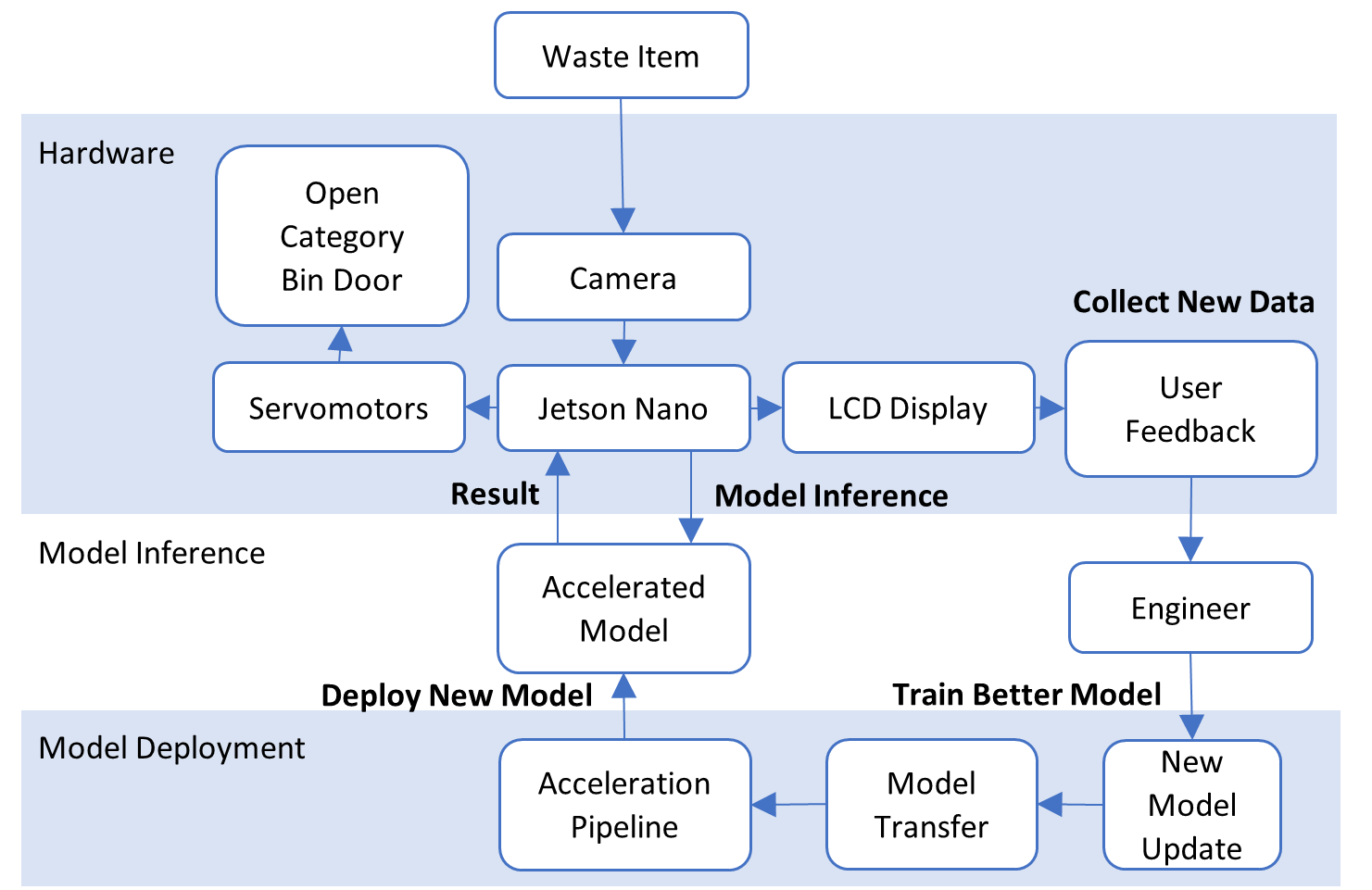}
        \caption{System block design of WasteNet.}
        \label{fig:sys_des}
\end{figure}
Figure \ref{fig:sys_des} shows the proposed internal system design of WasteNet and how the Jetson operates each component. The system is to be implemented in a SmartBin which is an artificially intelligent bin. There are 3 sections, hardware, model inference and model deployment. The hardware layer consists of the connections and components required to operate WasteNet such as the camera for input, servomotors to open the bin door and the display to show the user the result from classification. Due to the small training dataset of recycling images, we look to collect data from new inferences on new objects that would be stored locally. These would be sent to the engineer to retrain and deploy better versions of WasteNet. The model inference stage runs the accelerated model on new inputs, as well as handles power management such as when to enable inferencing. The final layer shows the model deployment stage in which new models are trained and accelerated for continuous deployment. This details the envisioned system design of WasteNet.

\subsection{Benchmark WasteNet Model}
The benchmark model chosen was WasteNet \cite{nine}, a recycling image classifier which classifies objects into 5 distinct categories: cardboard, glass, metal, paper, and plastic. The motivations for this have been discussed in Section 1.\\
In constructing the model, the chosen pre-trained base model, the model that we apply Transfer Learning to, is DenseNet169 as it was the closest model, in terms of parameter size, to the GoogLeNet base model used by \cite{nine} at 12.6M parameters. Both were also traditional convolutional networks in that they scaled in only one dimension (depth).\par
Transfer learning (TL) is a partial by-solution to reducing end-to-end model development times. Training a model from scratch is very long and unintuitive and so we instead take pre-trained base models that have already been trained on a certain dataset, for example ImageNet \cite{t1}, and train our target dataset on top of these pre-trained models [29-30]. The success of TL is highly dependent on the similarities between the pre-trained model domain and the target dataset domain \cite{t2} but is effective in producing high accuracy models.\par
A dataset's domain comprises of the spatial representation of the dataset's feature distributions. This could be a condensed feature map for the samples or could even be the general category by which the samples fall under. 
Many variations in image parameters such as lighting, colour maps, edges or gradients in images can change the domain or distribution of a sample dramatically. Two similar domains will yield better training and model results through transfer learning as the weights the model learns do not change drastically.

\subsection{EfficientNet WasteNet Model}
EfficientNets, a family of efficient CNNs, wildly outperform traditional CNNs on the ImageNet dataset all while containing less parameters and requiring a lower amount of floating-point operations \cite{eleven}. The aim of this paper is to implement the WasteNet network at the edge in the least expensive configuration that also outputs the highest accuracies on the test sets thus EfficientNets make the best base model for transfer learning. \par
Deep learning models are inherently feature extractors that are layered with different sized convolutional blocks to generate feature map representations of the input. Due to this we can use a pre-trained EfficientNet base that has already learned features of everyday objects from the ImageNet dataset and update its weights to tailor to the domain of recycling objects.\par
\begin{figure}[H]
        \centering
        \includegraphics[width=0.55\textwidth]{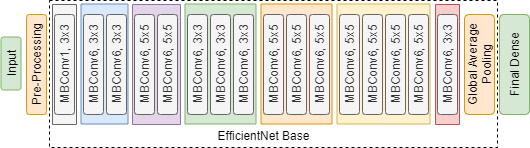}
        \caption{Proposed WasteNet Model Architecture.}
        \label{fig:arch}
\end{figure}
Figure \ref{fig:arch} shows the Efficient WasteNet model architecture. The architecture concatenates a pre-processing layer to apply data augmentation in attempts to avoid the model overfitting, a global average pooling layer to flatten the output feature map and finally a softmax activated dense layer for classification. The model layers are left unfrozen throughout training such that the model weights update accordingly to adapt to the new classification domain. There lies issues with 'catastrophic forgetting' when using transfer learning as the model weights may update too drastically causing the model to 'forget' its learned features. This is avoided through use of a low learning rate (LR), identified through a learning rate finder, to avoid large changes in these layer weights. The LR finder identifies a LR which leads to the lowest loss and should be configured within this low LR range to prevent 'forgetting'.
\par

\subsection{TensorRT Acceleration Pipeline}
TensorRT is NVIDIA’s programmable inference accelerating library which offers high-performance neural network inference optimisation to deliver low latency, high-throughput, power, and memory efficient runtime engines \cite{t8}.\par
\subsubsection{Quantisation and pruning in model acceleration}
TensorRT configurations include the use of quantisation and pruning, two key techniques in model acceleration. Quantisation and pruning are methods used in complexity and memory reduction of a network: Quantisation is the process of constraining values of a network from a continuous set or denser domain to a relatively discrete set and sparser domain \cite{t3}\cite{t6}. A form of this would be reducing the precision of the floating-point domain representation of weights of layers of the network. Typically, quantisation may be applied to the training or inference step of a network. Quantisation at the training step is typically found to severely impact the final accuracy for predicting correctly, known as the Top-1\% accuracy, of the model \cite{t3}. Instead quantisation is applied at the inference stage, and this is important for real-time applications as the reduction of operations and multiply-accumulate cycles (MAC) means for less processing to take place at inference \cite{t4}. Typical quantisation takes the form of 16-bit “float16” and 8-bit “INT8”.\par
Pruning is the process of removing weights from the network that fall below a certain threshold \cite{t6}. Many weights of the network may be close to zero and so would not contribute to the final classification. Pruning can then be used to remove these extra weights thereby reducing storage requirements of the network \cite{t5}. In both strategies’ quantisation will almost always be used when moving from computers to microprocessors. For instance, the 32-bit GPU of the Jetson Nano operates at 235.8GFLOPS, whereas operating on 16-bit values doubles its performance to 471.6GFLOPS, therefore it is desirable to constrict the model to 16-bit values.

\subsubsection{TensorRT Configurations}
\begin{figure}[H]
        \centering
        \includegraphics[width=0.5\textwidth]{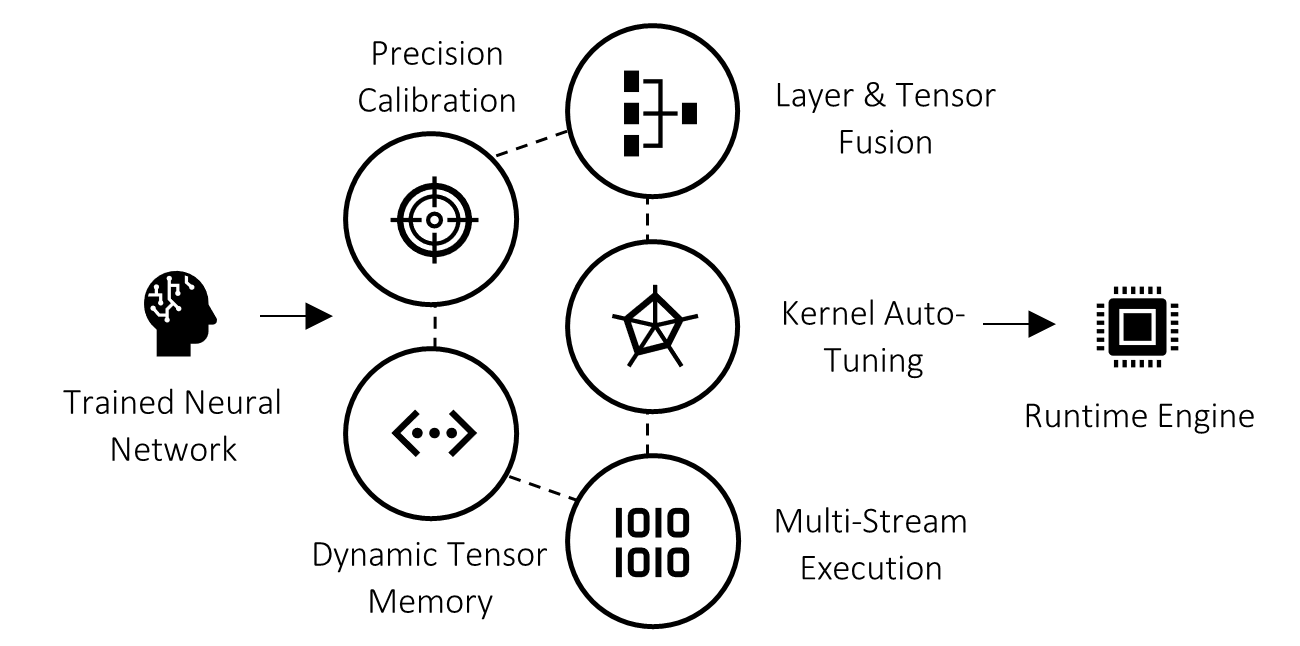}
        \caption{TensorRT optimisation pipeline (in no particular order).}
        \label{fig:trt}
\end{figure}

A runtime engine is the product of the library after a network has undergone the optimisation pipeline and this engine is what is used to run inference on the target device. The engine is particularly useful for embedded devices as it can be serialised and de-serialised, thereby allowing the loading of the optimised network without the need to build the engine at runtime which can be very costly. TensorRT integrates well with NVIDIA GPU’s due to being built on NVIDIA’s parallel programming model, CUDA. CUDA cores in NVIDIA GPU’s are specialised cores dedicated to increasing performance in important machine learning tasks such as forward and backward convolution, pooling, normalisation, and activation layers \cite{seven}. TensorRT then executes various optimisations and converts the neural network model into multiple graph forms that are run in parallel to allow for inference speed-ups of the model on these CUDA cores. Such optimisations are controlled at the user-level through conversion parametrisations. The user can also impose various other limitations such as setting the maximum memory allowed to be consumed by runtime engines, maximum batch sizes allowed through the engine and how the engine should be instantiated at runtime. Quantisation is also offered through precision calibration. The TensorRT pipeline is shown in Figure \ref{fig:trt}.\par
The optimiser optimises the use of GPU memory and bandwidth by fusing nodes in a kernel \cite{t8}. This means layers that take the same source tensors as inputs are fused together to give a single larger layer that performs the same arithmetic operations on the input tensor, with similar parameters, which increases sparsity and throughput of the network. The runtime engine also allows dynamic allocation of tensor memory buffers and minimises the networks memory footprint by re-using these buffers during inference calls made by the network. \par
Most notably, the main optimisation that the user can apply is the auto-tuning of kernels in the network’s layers. This occurs during the engine build phase where TensorRT eliminates layers in the network whose outputs are not used during inference, to avoid unnecessary computation. This is similar in nature to pruning the network. Then, where appropriate, convolution, ReLU and bias layers are fused to form a single layer and called at once. This improves the efficiency, the amount of throughput delivered per unit-power, of the network. This task is specifically tailored to the target devices GPU. The user can therefore drastically accelerate neural networks on NVIDIA GPU’s. Within this paper we apply these optimisations through various configuration to gauge the effectively on model acceleration on the Jetson Nano board.

\subsection{Connecting Servo-Motors for Latency Analysis}
In this paper we also analyse the Jetson Nano’s capability to drive servomotors and measure the latency from classification to actuator movement to gain insight into what hardware delays of the model. This is required as the final SmartBin design will have compartments which must open when the object is shown to the network and so we wish to verify real-time latency in a practical deployment.\par
To ensure an excess of calls are not sent to the servomotors due to the high IPS rate of the models, a queuing system is implemented in which the signal sent to the servo is only sent if the average result over the last group of inferences changed state. This queuing system also acts as a majority-vote classification system to negate random classification glitching during inference. This functionality means that if an object is shown to the camera input for a long time, the signal will not be repeatedly sent which may cause a backlog of actuator movement during execution.

\section{Methodology}
\subsection{Model Reconstruction Methodology}
\subsubsection{Benchmark Model}
The hyperparameters for augmentation of the input training set were kept the same from \cite{nine} and are shown in Table \ref{tab:benchpara}. The best train/validation/test splits were 72/18/10 with the 10\% test set kept the same from \cite{nine} to facilitate a fair evaluation of the end accuracy of the models. This split was identified from a 5-Fold Cross Validation split with 10\% of the dataset held out for the test set.  The validation split is set to 20\% of the remaining 80\% of the TrashNet set (18\%). Finally, a learning rate (LR) of 2e-5 was applied. \cite{nine} identified this learning rate through a learning rate finder which runs training of the model at different LR’s and finds which LR’s produce the least amount of loss after a few epochs of training.\par
The training procedure was also kept the same and saw 75 epochs at 2e-5 of the frozen model, and gradual unfreezing of layers of the base model while applying 15 epochs at 1e-6. We choose a lower learning rate for the unfrozen layers as to avoid catastrophic forgetting by over-training the model to the point where the pre-trained weights change too drastically. All training was carried out on Google Colab on a laptop, utilising their cloud GPU technology.\par

\begin{table}[H]
  \centering
    \small
    \caption{Benchmark model training and pre-processing parameter values.}
  \begin{tabular}{ll}
    \toprule
    \cmidrule(r){1-2}
    Pre-Processing Method     & Value \\
    \midrule
    Random Flip & Horizontal and Vertical \\
    Random Rotation & Up to 180\(^{\circ}\) \\
    Random Translation & Up to 10\% \\
    Random Zoom & Up to 100\% \\
    \bottomrule
    \cmidrule(r){1-2}
    Training Parameter     & Value \\
    \midrule
    Learning rate scheduler & Constant learning rate scheduler \\
    Optimizer & Adam optimizer \\
    Stopping criteria & 75 epochs \\
    Stopping criteria fine-tuning & 15 epochs \\
    Loss functions & Sparse categorical cross entropy \\
    Classifier activation function & Softmax \\
    Batch size & 16 \\
    Learning rate & 2e-05 \\
    Learning rate fine-tuning & 1e-06\\
    \bottomrule
  \end{tabular}
  \label{tab:benchpara}
\end{table}
The benchmark model was successfully recreated as per \cite{nine} and the exact test-set accuracy of 95.4\% was achieved on a desktop PC. The next step in this stage was to collect the benchmark performance metrics by running inference on the Jetson Nano with the benchmark model to establish a comparative baseline on the embedded device. However, the DenseNet model at 81 MB is too large to load onto the Jetson Nano and would cause the Nano to hang due to Out of Memory (OOM) error. Upon retrieval of the console and Tegrastrats logs it is apparent the model required much more than the available 4GB shared memory and would not work with any of the compression techniques to start running on the Nano. 
This was an anticipated result as the original WasteNet was not designed with the intended host device to be an embedded device like the Nano. It is apparent that for edge deployment these over-parameterised models will not suffice and that reconstruction using a mobile-friendly network architecture like that of EfficientNets is required.

\subsubsection{Optimising WasteNet for Mobile Architectures}
In this stage we begin construction of the optimised WasteNet model that will be implemented on the Jetson Nano. For this case, we begin to test our hypothesis that the EfficientNet base can be used to recreate a less resource intensive WasteNet model and that the model maintains its accuracy in comparison to the benchmark model.\par
The set of optimised models developed used B3, B2 and B0 from the EfficientNet family as base models to apply transfer learning to. B0 is the main focus as it approximately exhibits a similar ImageNet Top-1 accuracy to DenseNet169 (benchmark base) of around 77.2\%. The larger B3 and B2 are used to analyse if there is a parameter-accuracy trade-off over B0. \par
All models were trained using the same pre-processing and procedure as the benchmark model. The learning rate of 2e-5 of the benchmark model did not appear suitable for the EfficientNet base as the models took over 70 epochs to converge and so, we borrow the idea of utilising a learning rate finder from \cite{nine} set in the range of learning rates less than 1e-4. The quickest convergence of the model occurred at 4.3e-5 in which the model would converge to 90\% validation accuracy within a few epochs. The augmented training parameters are showin in Table \ref{tab:effpara}
\begin{table}[H]
  \centering
    \small
    \caption{EfficientNet optimised training parameter values.}
  \begin{tabular}{ll}
    \toprule
    \cmidrule(r){1-2}
    Training Parameter     & Value \\
    \midrule
    Learning rate scheduler & Constant learning rate scheduler \\
    Optimizer & Adam optimizer \\
    Stopping criteria & 50 epochs \\
    Stopping criteria fine-tuning & 8 epochs \\
    Loss functions & Sparse categorical cross entropy \\
    Classifier activation function & Softmax \\
    Batch size & 16 \\
    Learning rate & 4.3e-05 \\
    Learning rate fine-tuning & 4e-06\\
    \bottomrule
  \end{tabular}
  \label{tab:effpara}
\end{table}
A further training technique developed came from the fine-tuning documentation of TensorFlow \cite{f1} and \cite{twfour}. The idea of freezing and unfreezing layers when training had proved to increase the final accuracy and stability of the network in many literatures including the original model. The EfficientNet models were volatile in that they underfit slightly due to much larger validation accuracy compared to the training accuracy. \par
The idea of “consolidation epochs” was developed from the idea of fine-tuning but the base model layers are never frozen at any stage of the networks training cycle. With EfficientNet, model validation accuracy converges to the accuracy bottleneck within only a few epochs, fine-tuning only the top layers would have little impact on model performance. Instead, the entire model is trained at a lower learning rate to allow the training accuracy to catch up and prevent underfitting, all while keeping the network from not overtraining and forgetting thereby consolidating itself. \par 
The consolidation epochs would carry out a short “post-training” training cycle of around 8 epochs at an order of magnitude smaller LR (4e-6). This proved significantly important in stabilising the network output.\par
The new EfficientNet models were ran on the Jetson Nano and their performance metrics recorded. The models must be saved and loaded without their respective optimisers. This is because the optimiser holds the training status of the model, however, we do not require this and omitting the optimiser decreases model size and inference times.

\subsection{Model Acceleration Methodology}
\subsubsection{Intuitive Hyperparameter Analysis}
In this section we borrow ideas of compound scaling from the EfficientNet paper \cite{eleven} to intuitively fine-tune the previously developed B0 model and identify the boundary hyper parameterisation that outputs the highest inferencing speed pre-acceleration with no drop in classification accuracy. The time taken to capture the real-time inputs via the Raspberry Pi Camera Module on the Jetson Nano is around 20ms. This should be accounted for ahead of the acceleration stage and decreasing the capture resolution should lead to better inferencing speeds.\par
By downscaling the resolution of the inputs to the network more memory is then available in the shared memory of the Jetson Nano that can then be used for optimisation by TensorRT at runtime. This requires a simple re-training of the network in accordance with the same training procedure as mentioned in Section 4.1.2, but with the new downscaled resolutions. This was done with 25\% and 50\% reduction in resolution.TrashNet yields inputs of 512x384 pixels and so the new resolutions after scaling would be 384x288 pixels and 256x192 pixels, respectively. 

\subsubsection{Deep Compression Optimisations via TensorRT}
TensorRT offers float32 and float16 quantisation, alongside memory overhead and batch size limiting. Further to this, optimisations tailored to NVIDIA GPUs are applied which can accelerate models drastically.\par
We input the model through TensorRT and produce float32 and float16 runtime-engines to evaluate the speed differences between the models and verify GPU usage. The batch size for each configuration is set to 1 as we are running single input inferences and the max workspace size is set to 33MB (equivalent to string "1<<25" in code) as this allowed the model to run at near overhead memory capacity. This is the maximum amount of scratch space given to each layer in the model.\par
The accuracies of these runtime-engines are also tested using the same test-set as the original model. These accuracy tests are run on the Nano as opposed to on the PC through Colab as all runtime optimisations must be carried out on the device that will carry out the inference. This means the subgraphs will be created on the Jetson Nano. 
This allows for the library to pull all the overheads associated with the target device’s hardware and better optimise for memory and cache usages. It is noted that only models developed using TensorFlow 2.3 work with TensorRT and therefore downgrade of TF versions will be required.

\subsection{Evaluating Real World Model Performance and Model Generalisation Capabilities}
Unlike many papers that look to test state-of-the-art on a held-out test-set, few test the performance of the model on real world scenarios with real object inference. This section focuses on evaluating the generalisation capabilities of the model once deployed through a real world accuracy test and solutions to issues identified with the model within these tests.

\subsubsection{Real World Accuracy Test}
The real world accuracy test is different to running an accuracy test on a test-set during training, in that we are not testing the model on these ideal examples from the training dataset, but instead on the non-ideal inputs that the model would experience on a day-to-day basis after deployment to evaluate model generalisation capabilities. \par
\begin{figure}[H]
        \centering
        \includegraphics[width=0.3\textwidth]{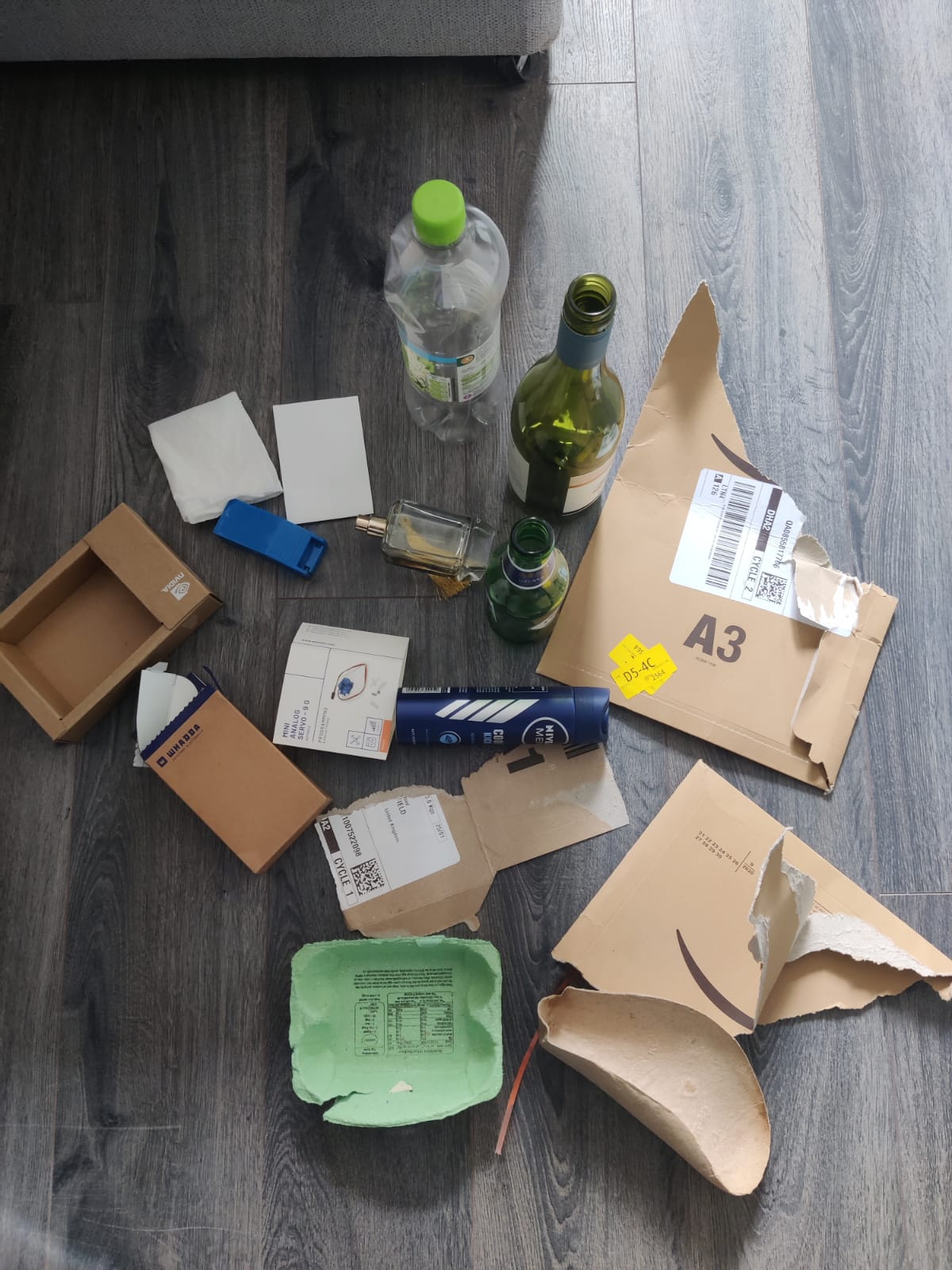}
        \caption{Sample of items used in real world accuracy test.}
        \label{fig:items}
\end{figure}

 
The test saw a collection of 40 recycling items that were taken from recycling bins for a true test of the type of recycling people throw out. The item type breakdown consisted of 9 Cardboard, 5 Glass, 6 Metal, 10 Paper and 10 Plastic samples. The camera was pointed at white sheet of A1 card to act as a neutral background to the objects during inference. A sample of the 40 items is shown in Figure \ref{fig:items}.\par
\begin{figure}[H]
        \centering
        \includegraphics[width=0.45\textwidth]{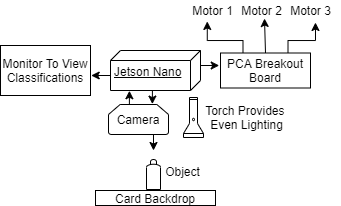}
        \caption{Block diagram of test-bed.}
        \label{fig:setupblock}
\end{figure}
\begin{figure}[H]
        \centering
        \includegraphics[width=0.35\textwidth]{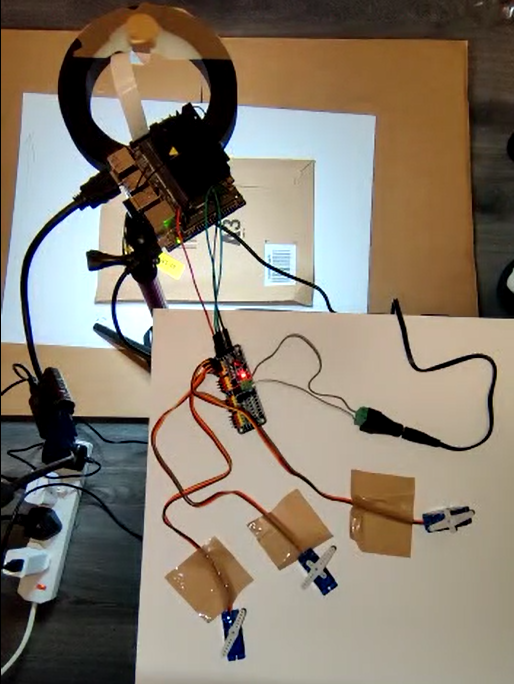}
        \caption{Test-bed setup using Jetson Nano. The Jetson Nano and its camera are mounted to a ring light. The test object is placed on a white A1 piece of card. We use a computer monitor as the display, connected via HDMI to the Jetson Nano. The servomotors are connected via a PCA breakout board and mounted to a table.}
        \label{fig:setup}
\end{figure}

The block diagram of the setup for real world inference is given in Figure \ref{fig:setupblock}. The Jetson Nano was connected to a monitor and sits on top of a ring-light. The ring light was set up above the centre of a card backdrop to provide even lighting across the object in question. A RaspberryPi V2 camera is connected to the Jetson Nano and is taped to the middle of the ring-light ring such that its field of view is filled by the white card. The servomotors are also connected to the Nano and taped individually on a surface as to be easily recorded for latency analysis. The Jetson Nano test-bed setup is shown in Figure \ref{fig:setup}.
\subsubsection{Mixed Dataset Training Technique}
In image classification tasks many models tend to completely fail at adapting to the domain shift that the real world imposed on inputs to models. This is because in real life the model classifies on unseen inputs in unfavourable lighting and positional conditions. Models are typically trained on “clean” datasets, that is datasets that have been crafted to be ideal to achieve maximum model accuracy and performance in a controlled test environment but this can lead to generalisation issues when the model is deployed due to overfitting of the model to unrealistic input standards of the training set.\par
\begin{figure}[H]
        \centering
        \includegraphics[width=0.25\textwidth]{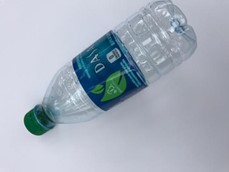}
        \includegraphics[width=0.14\textwidth]{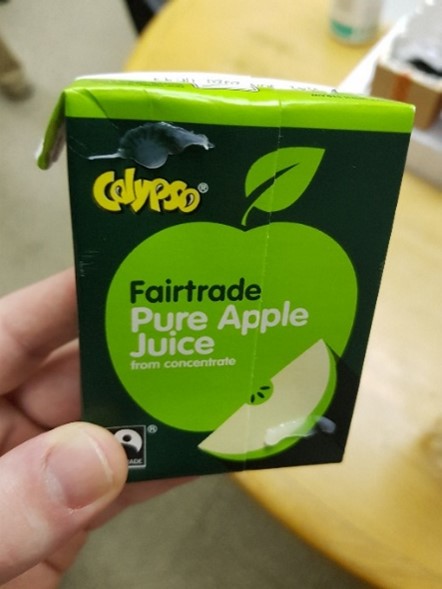}
        \caption{Sample image from the 'ideal' TrashNet dataset (left) versus sample image from the 'non-ideal' IBM dataset (right).}
        \label{fig:tnetex}
\end{figure}

To counteract overfitting, we can manually introduce a domain-shift to the training data ahead of model deployment to prevent the model closely learning the ideal dataset of the TrashNet. We introduce IBM’s trash dataset. This dataset is a private collection of recycling images constructed by the IBM WasteNet team. The IBM dataset consists of images of recycling in more realistic conditions with varying light and has non-ideal characteristics such as hands appearing in samples. It is considered a non-ideal dataset as it does not prioritise best training data practices \cite{f2}. An sample image from both datasets is shown in Figure \ref{fig:tnetex} for comparison.
\begin{table}[H]
 \caption{New mixed dataset model training parameters and suggested data augmentation parameters.}
  \centering
    \small
  \begin{tabular}{ll}
    \toprule
    \cmidrule(r){1-2}
    Pre-Processing Method     & Value \\
    \midrule
    Random Flip & Horizontal and Vertical \\
    Random Rotation & Up to 180$^{\circ}$ \\
    Random Translation & Up to 10\% \\
    Random Zoom & Up to 75\% \\
    \bottomrule
    \cmidrule(r){1-2}
    Training Parameter     & Value \\
    \midrule
    Learning rate scheduler & Constant learning rate scheduler \\
    Optimizer & Adam optimizer \\
    Stopping criteria & 45 epochs \\
    Stopping criteria fine-tuning & 30 epochs \\
    Loss functions & Sparse categorical cross entropy \\
    Classifier activation function & Softmax \\
    Batch size & 16 \\
    Learning rate & 4.3e-05 \\
    Learning rate fine-tuning & 4e-06\\
    \bottomrule
  \end{tabular}
  \label{tab:mixpara}
\end{table}
The IBM dataset was mixed with the TrashNet dataset in the same 72/18/10 split as used originally. The 10\% test-set was held out on its own and I introduce a new accuracy test for the IBM set separate from the TrashNet test-set. Also, a lower 75\% zoom augmentation limit was implemented into the data augmentation stage of training due to issues encountered with closeness of the subject to the camera in the optimised model real world accuracy test. The new models saw 45 epochs at 4.3e-5 in which the training accuracy stagnated. It took a further 30 consolidation epochs at 4e-6 to reach a validation accuracy. \par
A model trained purely on the IBM dataset was also created so that we could observe the accuracy distributions with unmixed and mixed training. Mixing datasets also serves to give more training examples for the model to learn on. The data augmentation and training parameters can be found in Table \ref{tab:mixpara}.

\subsection{Servo-Motor Latency and Board Power Analysis}
\subsubsection{Latency Analysis}
Latency analysis is carried out to identify the signal delay within the Jetson Nano from inferencing to actuator movement of the servo-motor. Video recordings were used in conjunction with the model IPS rate to calculate the exact timing for signal propagation and how that timing splits between the average queuing function delay and servo latency delay which is shown in Equation \ref{eqlat}.
\begin{equation}
    \label{eqlat}
    T_{total}=
    \left [ \frac{N^{*}}{2}+1 \right ]\cdot \frac{1}{IPS}+ \left (  T_{cts}-\left [ \frac{N}{2}+1 \right ]\cdot \frac{1}{IPS}\right )
\end{equation}
\(T_{total}=T_{Queue}+T_{Servo}\) such that \(N^{*}\) represents the desired queue length, IPS is the model throughput, \(T_{cts}\) is the classification-to-servo delay and N is the queue length used within the experiment which was set to 10 A higher queue size should lead to a more stable output as the averaging is calculated across more inferences but will lead to a higher delay in the the SmartBin door opening.

\subsubsection{Power Analysis}
The Jetson Nano is intended to be powered off a battery to meet the portability requirement. The Jetson Nano contains two power profile modes: a 5W low-power state and a 10W max-power state (MAXN). The average power from the Jetson Nano can be read from the Tegrastats, specifically the POM\_5V\_IN value which shows values X/Y, X being the current drawn power and Y being the average drawn power by the board. The average power is measured for the Jetson in an idle state with no model loaded, then in an idle state with a model loaded and finally while the model is inferencing for both the 5W and 10W power profiles.


\begin{table*}[!ht]
 \caption{Comparison of results between the benchmark model and EfficientNet models.}
  \centering
  \begin{adjustbox}{width=1\textwidth}
    \small
  \begin{tabular}{lllllllll}
    \toprule
    \cmidrule(r){1-2}
    \multicolumn{1}{p{2cm}}{\centering Model \\ Architecture}     & \multicolumn{1}{p{2cm}}{\centering Total \\ Parameters}     & \multicolumn{1}{p{2cm}}{\centering Model \\ Size (MB)}     & \multicolumn{1}{p{2cm}}{\centering Decrease in Model Size from Benchmark (\%)}     & \multicolumn{1}{p{2cm}}{\centering Total \\ Epochs}     & \multicolumn{1}{p{2cm}}{\centering Top-1 \\ Validation\\Accuracy (\%)}     & \multicolumn{1}{p{2cm}}{\centering Top-1 \\ Accuracy on\\Test Set (\%)} & \multicolumn{1}{p{2cm}}{\centering Top-1 \\ Accuracy on\\Test Set\\ on Jetson (\%)} & \multicolumn{1}{p{2cm}}{\centering Inference \\ Time on \\ Jetson (s)} \\
    \midrule
    \multicolumn{1}{p{2cm}}{\centering DenseNet169} & \multicolumn{1}{p{2cm}}{\centering 12642880}  & \multicolumn{1}{p{2cm}}{\centering 81}  & \multicolumn{1}{p{2cm}}{\centering 0}  & \multicolumn{1}{p{2cm}}{\centering 91}  & \multicolumn{1}{p{2cm}}{\centering 95.7}  & \multicolumn{1}{p{2cm}}{\centering 95.3781}  & \multicolumn{1}{p{2cm}}{\centering Out of memory}  & \multicolumn{1}{p{2cm}}{\centering -}     \\
    \midrule
    \multicolumn{1}{p{2cm}}{\centering EfficientNetB3} & \multicolumn{1}{p{2cm}}{\centering 10783535}  & \multicolumn{1}{p{2cm}}{\centering 53}  & \multicolumn{1}{p{2cm}}{\centering 14}  & \multicolumn{1}{p{2cm}}{\centering 50}  & \multicolumn{1}{p{2cm}}{\centering 97.9}  & \multicolumn{1}{p{2cm}}{\centering 95.7983}  & \multicolumn{1}{p{2cm}}{\centering Out of memory}  & \multicolumn{1}{p{2cm}}{\centering -}     \\
    \midrule
    \multicolumn{1}{p{2cm}}{\centering EfficientNetB2} & \multicolumn{1}{p{2cm}}{\centering 7768569}  & \multicolumn{1}{p{2cm}}{\centering 30}  & \multicolumn{1}{p{2cm}}{\centering 62.3}  & \multicolumn{1}{p{2cm}}{\centering 48}  & \multicolumn{1}{p{2cm}}{\centering 97.3}  & \multicolumn{1}{p{2cm}}{\centering 95.7983}  &
    \multicolumn{1}{p{2cm}}{\centering 95.7983} & \multicolumn{1}{p{2cm}}{\centering 32} \\
    \midrule
    \multicolumn{1}{p{2cm}}{\centering EfficientNetB0} & \multicolumn{1}{p{2cm}}{\centering 4049571}  & \multicolumn{1}{p{2cm}}{\centering 15.3}  & \multicolumn{1}{p{2cm}}{\centering 81.1}  & \multicolumn{1}{p{2cm}}{\centering 41}  & \multicolumn{1}{p{2cm}}{\centering 96.5}  & \multicolumn{1}{p{2cm}}{\centering 95.3781}  &
    \multicolumn{1}{p{2cm}}{\centering 95.3781} & \multicolumn{1}{p{2cm}}{\centering 0.32} \\
    \midrule
    \multicolumn{1}{p{2cm}}{\centering EfficientNetB0 (75\% Resolution)} & \multicolumn{1}{p{2cm}}{\centering 4049571}  & \multicolumn{1}{p{2cm}}{\centering 15.3}  & \multicolumn{1}{p{2cm}}{\centering 81.1}  & \multicolumn{1}{p{2cm}}{\centering 41}  & \multicolumn{1}{p{2cm}}{\centering 96.5}  & \multicolumn{1}{p{2cm}}{\centering 95.3781}  &
    \multicolumn{1}{p{2cm}}{\centering 95.3781} & \multicolumn{1}{p{2cm}}{\centering 0.15} \\
    \midrule
    \multicolumn{1}{p{2cm}}{\centering EfficientNetB0 (50\% Resolution)} & \multicolumn{1}{p{2cm}}{\centering 4049571}  & \multicolumn{1}{p{2cm}}{\centering 15.3}  & \multicolumn{1}{p{2cm}}{\centering 81.1}  & \multicolumn{1}{p{2cm}}{\centering 41}  & \multicolumn{1}{p{2cm}}{\centering 94.3}  & \multicolumn{1}{p{2cm}}{\centering 92.2818}  & \multicolumn{1}{p{2cm}}{\centering Not tested due to accuracy drop}  & \multicolumn{1}{p{2cm}}{\centering -}\\
    \bottomrule
  \end{tabular}
  \end{adjustbox}
  \label{tab:finalresults}
\end{table*}

\section{Results}
\begin{figure}[H]
  \centering
  \includegraphics[width=0.5\textwidth]{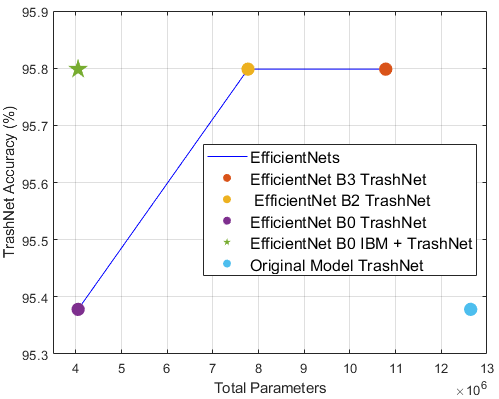}
  \caption{All model accuracy scores vs. model parameterisation. The green star indicates the final developed mixed-dataset model.}
  \label{fig:finalgraph}
\end{figure}

\subsection{Overall Summary of Results}
A plot showing each key development stage and their associated model constructed is shown in Figure \ref{fig:finalgraph}. For this plot we wish to be towards the upper-left quadrant, and we note the difference between the original WasteNet model, denoted by the cyan dot, and the new, much lower parameterised models with accuracies equal to and above the original model. There is an apparent accuracy bottleneck of 96\% across all models and this is because there is not enough training data for the models. We have around 2600 training data samples which is very small – a rule of thumb is we require around 1000 images per class \cite{f3}, so we should have around 5000 images, double the current dataset. This can be corrected in the future through continual learning, whereby we collect new samples from the users of the SmartBin. The experiments also showed the positive influence that a bigger and more realistic conditioned training dataset has on a model’s performance in the real world.\par
The biggest issues involved how far the object was from the camera input. We saw that zoom augmentation during the train phase led to these distance issues, and that reducing the maximum zoom to around 75\% helped fix this. This suggests the use of data augmentation techniques such as zoom build a discrepancy between actual objects seen by the classifier at testing than what is taught to the classifier during training. This is because a 2D image zoom conducted during pre-processing is not entirely the same as bringing the camera sensor closer to the object in real-life. 
In real life there is typically field-of-view and skew shifts which cannot be fully mimicked by 2D post zoom. Fixing this may lead to the model performing much better in the real world.\par
Table \ref{tab:finalresults} depicts a detailed account for each experiment and its respective results and analysis.

\subsection{Model Reconstruction Results}
The results in Table \ref{tab:finalresults} show that the EfficientNets, at a much lower parameter and model size, perform just as well, if not better, as WasteNet base models than the original DenseNet variant. The B3 and B2 WasteNet variants achieved a higher test-set accuracy of 95.8\% than the benchmark model, yet both threw OOM errors. The B2 model would load but the entire model could not be tended to on shared memory due to OOM error when allocating memory buffers corresponding to the outputs of the network. This can be solved by creating a 4GB swap file on the Nano’s ZRAM in which the final tensors of the B2 model can be loaded to, however, access to the ZRAM is much slower than that of the shared memory of the CPU/GPU and therefore the B2 model exhibited significantly slower inference times than what would be expected. The B0 model was small enough to have all its input and output tensors allocated to the shared memory and did not throw OOM errors. The un-optimised model ran at 320ms inference time per pass (3 Inference Per Second) on the Nano and maintained its accuracy.\par
The results of B0 provide evidence that EfficientNets are great transfer-learning bases for mobile architectures as the same model accuracy was achieved as the benchmark model at 8 million less parameters and the model was successfully loaded onto the Jetson Nano. However, the IPS is too low for any real-time application use.

\subsubsection{Resolution Drop}
When identifying maximal hyperparameter boundaries for input shapes, Table \ref{tab:finalresults} shows that at 50\% resolution scaling, we experience a drop in the Top-1 test-set accuracy due to the input being too small for effective training, but at 75\% scaling the model retains its accuracy. We also see a reduction of around 17ms in the inferencing time for the model at 50\%. This is then able to offset the increase in inferencing time when using the camera module. 25\% downscaling would then be the maximum boundary scaling factor for pre-acceleration.\par
A side observation to note is that the convergence of loss to its minimal during training occured much faster at the 50\% scaling factor. This is indicative that we can train larger models faster through variable resolution scaling where we can scale inputs progressively such that they learn and converge quickly but to a lower accuracy, and then can be scaled up accordingly to allow better features to be learnt and model accuracy to increase. This is beyond the scope of this paper but is future work for a larger dataset optimised WasteNet model.

\subsection{Model Acceleration Results}
\begin{figure}[H]
  \centering
  \includegraphics[width=0.49\textwidth]{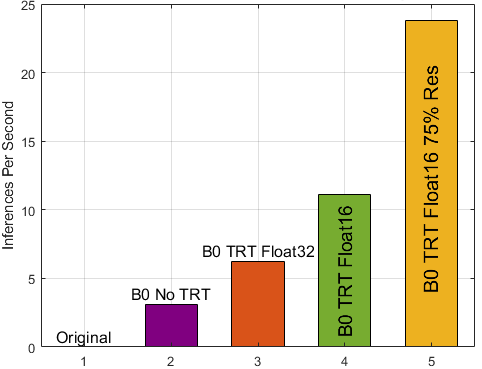}
  \caption{Model inference speeds (IPS) for each acceleration configuration.}
  \label{fig:trtgraph}
\end{figure}

\begin{table}[H]
 \caption{Model performance metrics for different TensorRT acceleration configurations.}
  \centering
  \begin{adjustbox}{width=0.49\textwidth}
    \small
  \begin{tabular}{llll}
    \toprule
    \cmidrule(r){1-2}
    \multicolumn{1}{p{2cm}}{\centering Model \\ Architecture}     & \multicolumn{1}{p{2cm}}{\centering Model \\ Size (MB)}     & \multicolumn{1}{p{2cm}}{\centering Average Inference Time (ms)}     & \multicolumn{1}{p{2cm}}{\centering Top-1 \\ Accuracy on\\Test Set (\%)} \\
    \midrule
    \multicolumn{1}{p{2cm}}{\centering EfficientNetB0} & \multicolumn{1}{p{2cm}}{\centering 15.3}  & \multicolumn{1}{p{2cm}}{\centering 320}  & \multicolumn{1}{p{2cm}}{\centering 95.3781}\\
    \midrule
    \multicolumn{1}{p{2cm}}{\centering TensorRT B0 @FP32} & \multicolumn{1}{p{2cm}}{\centering 38.1}  & \multicolumn{1}{p{2cm}}{\centering 160}  & \multicolumn{1}{p{2cm}}{\centering 95.3781} \\
    \midrule
    \multicolumn{1}{p{2cm}}{\centering TensorRT B0 @FP16} & \multicolumn{1}{p{2cm}}{\centering 49.5}  & \multicolumn{1}{p{2cm}}{\centering 90}  & \multicolumn{1}{p{2cm}}{\centering 95.3781} \\
    \midrule
    \multicolumn{1}{p{2cm}}{\centering TensorRT B0 @FP16 @75\% Resolution} & \multicolumn{1}{p{2cm}}{\centering 49.3}  & \multicolumn{1}{p{2cm}}{\centering 43}  & \multicolumn{1}{p{2cm}}{\centering 95.3781}
    \\
    \bottomrule
  \end{tabular}
  \end{adjustbox}
  \label{tab:trtresults}
\end{table}
The reconstruction results revealed B0 to be the best variant for the WasteNet model and so all results henceforth concern the B0 variant only.\par
The results in Table \ref{tab:trtresults} also show an increase of model sizes across configurations. This is expected as more compatible subgraphs would be created for the quantised configurations due to their truncated value nature, which leads to an increasing trend of model sizes across the variants. The increase over the default model is due to the optimised variants containing runtime engines, which we set to be pre-built. To pre-build the engines, sample inferences must be passed to the acceleration script of equal input shapes to that of the expected input shapes. By pre-building the runtime engines, we save resources by not having to build the engines from scratch at runtime/inference which can be extremely costly and slow down model load times drastically. This means a trade-off in model size for load times is present with TensorRT accelerations.\par
Utilising the set out configuration in Section 4.2.2, the results of Figure \ref{fig:trtgraph} show a trend of double inference speed increases between the default B0 model, the optimised Float32 model and the optimised Float16 model, while maintaining model accuracy across all variants. The Jetson Nano GPU operates at double the FLOP rate when running on 16-bit values explaining why the Float16 configuration performed the best. The Float16 model achieved the 11IPS bringing the model to the 10IPS 'real-time' threshold indicated by our performance goals. Furthermore, applying the Float16 configuration to the 75\% resolution model achieved a much higher 24IPS, a 750\% speed increase over the non-accelerated model and infinitely faster than the original model that could not be loaded onto the Nano!\par

\subsection{Evaluating Real World Model Performance and Model Generalisation Capabilities}
\subsubsection{Original Optimised Model Results}
\begin{figure}[H]
  \centering
  \includegraphics[width=0.49\textwidth]{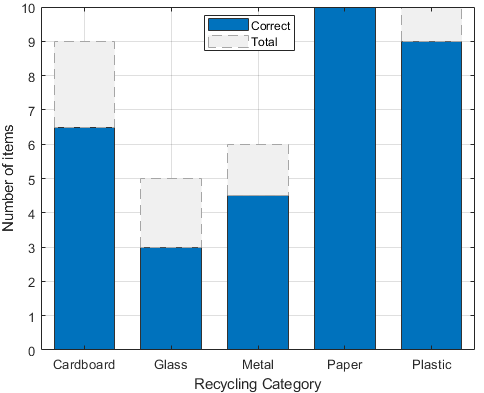}
  \caption{Bar graph of the mixed-dataset model's accuracy scores per category of recycling. The total objects collected for each category is indicated by the grey silhouette bars. The number of correct items predicted is given by the blue bars.}
  \label{fig:accgraph1}
\end{figure}

\begin{table}[H]
 \caption{Overall real world accuracy for the model.}
  \centering
  \begin{adjustbox}{width=0.4\textwidth}
    \small
  \begin{tabular}{lll}
    \toprule
    \cmidrule(r){1-2}
    \multicolumn{1}{p{2cm}}{\centering Model Architecture}     & \multicolumn{1}{p{2cm}}{\centering Accuracy Including Errors}     & \multicolumn{1}{p{2cm}}{\centering Accuracy Excluding Errors} \\
    \midrule
    \multicolumn{1}{p{2cm}}{\centering B0 @75\% Resolution} & \multicolumn{1}{p{2cm}}{\centering 82.5\%}  & \multicolumn{1}{p{2cm}}{\centering 90\%} \\
    \bottomrule
  \end{tabular}
  \end{adjustbox}
  \label{tab:acctable1}
\end{table}
For Table \ref{tab:acctable1}, 'Accuracy including errors' refers to the true accuracy of the model when including the identified errors such as the zoom and lighting problems. 'Accuracy excluding errors' represents the maximum theoretical accuracy of the model if these pertubations were resolved and serves to show the generalisation capabilities of the model in the real world. The results from Figure \ref{fig:accgraph1} shows the model scored an accuracy of 82.5\%. During the test it was observed that it was plagued with various issues when inferencing on the objects shown to it. The accuracy if these issues had not been present would be 90\% which is below the test-set model accuracy of 95.4\% predicted in the test-set accuracy test.\par
The model tended to confuse cardboard for paper and glass for metal. The prior occurred when the cardboard object being inferred on did not cover the entire field-of-view of the camera input. It could be suspected that the white card background used during inference was interfering with the cardboard when it did not cover the entire view of the camera, however if this were the issue then the other objects such as the metal cans which never cover the entire view of the camera should exhibit the same issue.\par 
When inferencing on glass and metal objects the camera view shows a glare due to the reflection of the overhead ring-light off of the shiny surface of the objects, leading to poor white balance and a very dim input into the model.\par Furthermore, all categories suffered from zoom issues. When objects such as the cardboard or glass bottles that would incorrectly be classified were brought closer to the camera view the model tended to correctly identify the objects. This zoom issue explains the problem with the cardboard being misclassified. The glare issue was fixed by orienting the ring-light at a 15° angle to the normal of the card backdrop. This off-axis tilt meant reflections from the objects did not hit the camera lens and led to correct classifications.\par
\begin{figure}[H]
  \centering
  \includegraphics[width=0.24\textwidth]{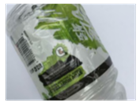}
  \includegraphics[width=0.24\textwidth]{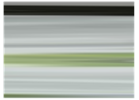}
  \caption{Training samples fed to the model after data augmentation using 100\% zoom.}
  \label{fig:zoomfig}
\end{figure}

Further investigation into the zoom issue discovered a flaw in the data augmentation techniques proposed by \cite{nine}. Using 100\% zoom to augment training samples led to very poor inputs to the model during training (see Figure \ref{fig:zoomfig}) and very high regularisation.\par
The 100\% zoom is too great a zoom and destroys true representation of the original object. Limiting zoom to a more sensible value of around 75\% proved to resolve some of the zoom issues experienced by the model. \par 
It is apparent the model is not robust enough to generalise well. This could be due to slight overfitting exhibited during the ‘consolidation epoch’ training stages as it leads to the model being unable to generalise to other objects.

\subsubsection{Mixed Dataset Training Results}
\begin{figure}[H]
  \centering
  \includegraphics[width=0.49\textwidth]{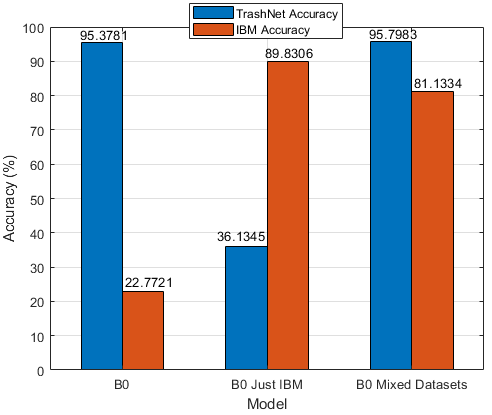}
  \caption{Accuracy of the new models on the TrashNet test-set and IBM test-set.}
  \label{fig:mix1}
\end{figure}

\begin{figure}[H]
  \centering
  \includegraphics[width=0.49\textwidth]{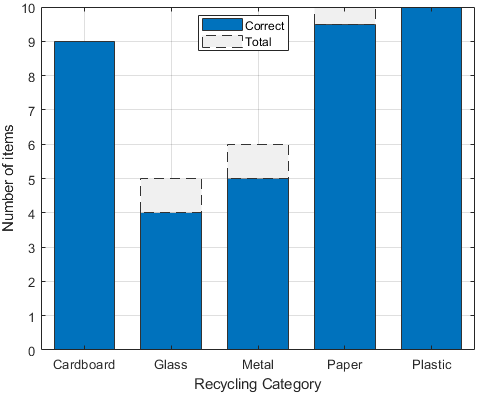}
  \caption{Bar graph of the mixed-dataset model's accuracy scores per category of recycling. The total objects collected for each category is indicated by the grey silhouette bars. The number of correct items predicted is given by the blue bars.}
  \label{fig:mix2}
\end{figure}

Generalisation capabilities of the models can be measured by the ability for a model to achieve sufficient accuracies on both ideal and non-ideal model inputs. As can be seen in Figure \ref{fig:mix1} the only model that achieves this is the B0 model trained on the mixture of the IBM and TrashNet datasets. Not only this but the model sets a higher test-set accuracy of 95.7983 on TrashNet, which was previously only achieved by the larger B2 and B3 models. This higher accuracy attained over the original B0 provides evidence of the positive effect of mixing datasets. However, to verify the extent to which this domain-shift impacts the robustness of the model in the real world, the 40-item real world test is used with this new mixed dataset model. The results for this can be seen in Figure \ref{fig:mix2} and Table \ref{tab:mixtab} respectively.\par
The new model scores a much higher accuracy 94\% with the camera at 15° to the normal to fix the glare issues. This is much more acceptable and provides evidence for less overfitting occurring as the model can generalise much better than before. Decreasing the maximum zoom during the data augmentation phase also removed the zoom issues that occurred in the previous model as expected. Contributions to this increased accuracy may be due to the increased dataset size due to mixing the datasets. 

\begin{table}[H]
 \caption{Overall real world accuracy for the model.}
  \centering
  \begin{adjustbox}{width=0.4\textwidth}
    \small
  \begin{tabular}{lll}
    \toprule
    \cmidrule(r){1-2}
    \multicolumn{1}{p{2cm}}{\centering Model Architecture}     & \multicolumn{1}{p{2cm}}{\centering Accuracy Including Errors}     & \multicolumn{1}{p{2cm}}{\centering Accuracy Excluding Errors} \\
    \midrule
    \multicolumn{1}{p{2cm}}{\centering B0 Mixed Dataset @75\% Resolution} & \multicolumn{1}{p{2cm}}{\centering 94\%}  & \multicolumn{1}{p{2cm}}{\centering 95\%} \\
    \bottomrule
  \end{tabular}
  \end{adjustbox}
  \label{tab:mixtab}
\end{table}

\subsection{Power Analysis}
We conduct power analysis as the target deployment within bins means it should ideally be portably powered and so we wish to gain an understanding of the power consumption of the board at different stages of model usage.\par
\begin{figure}[H]
  \centering
  \includegraphics[width=0.49\textwidth]{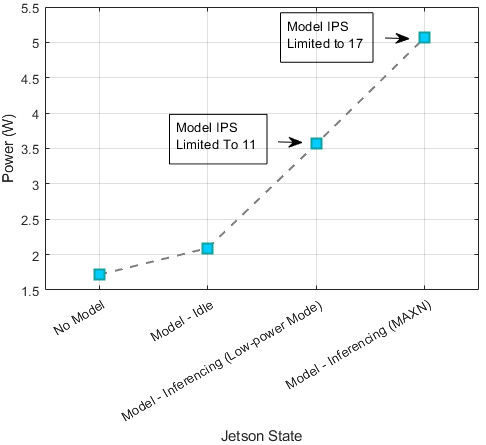}
  \caption{Total input power to the Jetson Nano per model state.}
  \label{fig:pow1}
\end{figure}

Figure \ref{fig:pow1} shows that sitting idle with the model loaded but not inferencing, the Jetson Board consumes a total of 2W which would require a 1400Wh capacity battery pack to operate for a month. For reference a typical mobile phone battery of around 5000mAh battery at 5V output gives around 25Wh. Therefore, sitting idle, the Jetson could be powered for 12.5 hours which is too little. Suggested solutions could be to use a much larger battery, however this can become costly and would render the cost restraint required by edge deployment redundant.\par
An interesting observation is that setting the Jetson to Low Power 5-Watt mode limits the model IPS rate from the standard 17IPS to 11IP but also decreases power usage of the model from 5W at inferencing to 3.5W at inferencing. This means an increased battery lifetime to 26 hours under constant inferencing. This decreased IPS rate is a trade-off but is not entirely bad for the model itself. 11IPS is still over the 10IPS real-time threshold set out in Section 1 and it would mean around one day's worth of constant inferencing. Battery life could be extended further by smartly switching on inference only when an object is sensed in front of the camera. Another route could be to just power the Jetson from a mains power supply, and this would still be a much more energy efficient option than using an entire PC through the benchmark model.

\begin{figure}[H]
  \centering
  \includegraphics[width=0.49\textwidth]{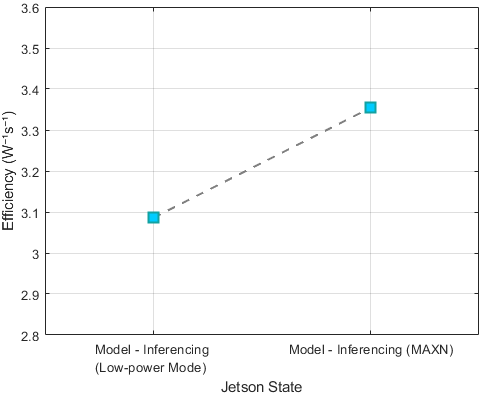}
  \caption{Model efficiency for each power profile.}
  \label{fig:eff1}
\end{figure}
The efficiency of our models is  the throughput (IPS) per unit power and is plotted in Figure \ref{fig:eff1}. The greatest efficiency occurs with the MAXN 10-Watt power profile at $3.6W^{-1}s^{-1}$, $0.8W^{-1}s^{-1}$ higher than the 5-Watt power profile. This suggests that the model should be set to inference at only the 10-Watt profile as this gives a greater use of input power to the board, however this result is contested by the need to prolong the Jetson Nano’s battery life to meet power demands. The power analysis concludes that for ease of use, the Jetson Nano should just be connected to the mains power supply. It is unfeasible to try to use a 1400Wh battery as they are very large and cost upwards of \$3000. This would limit commercialisation to only environments where power sockets are available such as offices or households.

\subsection{Latency Analysis}
The observed latency from which an object was shown to the camera to the correct servo motor rotating was found to be proportional to N, the size of the averaging queue function (see Equation 1). The actual latency of the model signal propagation within the Jetson Nano can be calculated as follows:\par
The time taken from showing the object to the servomotor moving was approximately 1 second (taken from the demo footage).\par
For the average queuing function which was set to size $N = 10$, the majority vote output will switch state once the new object occupies the majority of the queue which occurs after 6 inferences as $6/10$ of the states in the queue will correspond to the new object. In the demo the model was running at 17 IPS which is 59ms time to inference. Therefore,
\begin{equation*}
    6 *0.059s=0.354s
\end{equation*}
Now, for the total 1 second latency of the servo, we calculate the actual time for the servo to rotate from a signal being sent over the I2C channels from the second term in Equation 1,
\begin{equation*}
    Servo Latency=1s-0.354s=0.646s
\end{equation*}
so, the base latency to control the servo is around 0.65 seconds. The total expected latency for an increasing queue size N is then given by Equation \ref{eqlat2}.
\begin{equation}
    \label{eqlat2}
    T_{Total}=T_{Queue} + T_{Queue}=0.059\left [\frac{N}{2}+1  \right ] + 0.646
\end{equation}
A higher size N for the queue will give a more stable and reliable system output, as the majority vote holds more members or states, but leads to a decrease in system throughput as the servo delay increases proportionally. With the signal propagation delay found to be 0.646s, the system actuates the bin door open at a sixth of a second, much faster than what a human would take to classify and place their recycling into the bin.

\section{Concluded Outcomes}
The research conducted aimed to produce a methodology for optimising traditional ConvNet-based models and an acceleration pipeline for deploying these models to run at real-time at the edge on an NVIDIA Jetson Nano. The effects that real world deployment had on model accuracies was evaluated and solutions presented to increase generalisation capabilities of the model. A new, higher accuracy WasteNet model was developed with improved generalisation over the benchmark model of \cite{nine}. Further to this, an eco-system design was produced as we look to commercialise WasteNet as a product by conducting power and latency analysis on the board.\par
The acceleration pipeline first involves reconstruction of the benchmark model using EfficientNets. This stage requires domain analysis of the real world application domain to that of the data being trained on – it was identified that to maintain accuracy at deployment, it was not enough to train on ideal conditioned datasets that lead to the best test-set accuracies, but instead on mixed data that contains subjects in various exposures and orientations that mimic how they would be captured in the real world. For the Jetson Nano, model reconstruction utilised EfficientNetB0 as the base model to apply transfer learning to. The highest test-set accuracies were achieved using augmentation techniques such as flipping, up to 180° rotation and up to 75\% zooming. The ideal training parameters consisted of using a constant learning rate scheduler of 4.3e-5 learning rate, an Adam optimizer with sparse categorical cross entropy and a Sotmax activation function for the final classification layer. \par
To stabilise the model test-set accuracy a 'post-training' training cycle at a  learning rate of 4e-6 is required. This fine-tuning technique proposed called ‘Consolidation Epochs’ at a stopping criterion of 15 epochs slows learning to stabilise validation accuracies while preventing overfitting and this method proved vital in increasing the final accuracy of the model by up to 3\% post-training. \par
The maximum resolution reduction to the inputs of the model that would increase throughput while maintaining model accuracy was found to be 25\%. A higher test-set classification accuracy of 95.8\% was recorded for WasteNet. The best TensorRT configuration for maximum acceleration was identified to be 16-bit quantisation with a maximum workspace size of 33MB or "1<<25" bytes and pre-building the engine ahead of inferencing. The model inference rate increased from 3IPS to 23IPS achieving real-time rates. \par
Training the model on more realistic deployment conditions increased it's generalisation capabilities in the real world and led to an increase in the real world accuracy from 82.5\% to 94\%. A discrepancy was observed between zoom augmentation at training, and physically moving objects closer to the camera in the real world that led to a negative impact on classification accuracies. This result indicates more careful consideration is required when choosing data augmentation factors at training as these may lead to perception and scaling errors of objects at deployment.

\section{Future Work}

Future research can look to collect more dataset samples to increase the amount of data for a better distribution of training samples. Continual learning integration is a route for this, in which we obtain new training samples from inferences conducted by users and re-train the model on these. This would mean an ever-expanding dataset as there are many different variations of objects for a given recycling category and would relieve the training accuracy bottle neck. \par
An important recommendation is to look at fixing the train-test augmentation discrepancy through the work of \cite{twthree} introduced at the NeurIPS 2019 conference. This work proposes that fine-tuning the model post-training with samples of equal nature to that seen at deployment has a positive impact on decreasing this discrepancy.

\section{Acknowledgements}

The author pays his acknowledgments to the UCL Department of Electronic and Electrical Engineering and in particular Gerald McBearty and Andrew Moss for facilitating much of the equipment required for the work. I would also like to thank Mindy Yang and Gary Thung for collecting and providing the TrashNet dataset, as well as the IBM WasteNet team for allowing usage of their private recycling dataset for model development. 



\normalsize
\bibliography{tarticle}
\end{document}